\documentclass[runningheads]{llncs}
\usepackage[T1]{fontenc}
\usepackage[letterpaper,margin=1in]{geometry}
\usepackage{graphicx}
\usepackage{amsmath}
\usepackage{booktabs}
\usepackage{algorithm}
\usepackage{algpseudocode}
\usepackage[section]{placeins}
\usepackage{url}
\usepackage{color}
\usepackage[hidelinks]{hyperref}

\begin{document}
\title{Neuro-Symbolic Meta-Policies for Temporal Knowledge-Graph Memory under Partial Observability\thanks{This version of the contribution has been accepted for publication at ISWC 2026, after peer review, but is not the Version of Record and does not reflect post-acceptance improvements or any corrections. The Version of Record is available online at: \url{https://doi.org/10.1007/978-3-032-42015-2_10}. Use of this Accepted Version is subject to the publisher's Accepted Manuscript terms of use: \url{https://www.springernature.com/gp/open-research/policies/accepted-manuscript-terms}.}}
\titlerunning{Neuro-Symbolic Meta-Policies for Temporal KG Memory}
%
\author{Taewoon Kim\inst{1,2} \and Vincent Fran\c{c}ois-Lavet\inst{2} \and Michael Cochez\inst{3}}
\authorrunning{T. Kim et al.}
%
\institute{HumemAI, Amsterdam, The Netherlands\\
\email{taewoon@humem.ai} \and
Vrije Universiteit Amsterdam, Amsterdam, The Netherlands\\
\email{vincent.francoislavet@vu.nl} \and
ELLIS Institute Finland \& \AA{}bo Akademi University, Turku, Finland\\
\email{michael.cochez@abo.fi}}
\maketitle              
\begin{abstract}
Partially observable reinforcement learning requires deciding what to retain, retrieve,
and forget over time. We introduce a neuro-symbolic meta-policy that learns \emph{which}
symbolic memory heuristic to apply at each decision point while keeping execution
symbolic. Our setting uses temporal knowledge-graph memory in RoomKG, where hidden
state and observations are represented as Resource Description Framework (RDF) graphs
and memory is augmented with temporal RDF triple annotations. The model combines
knowledge-graph encoding of memory contents with value heads for question answering,
exploration, and forgetting, yielding a controller that is both adaptive and
inspectable. This gives the work a direct Semantic Web grounding through RDF-based
representation, annotation-compatible graph semantics, and graph-based symbolic
operations over explicit memory state. On train/test room splits at long-term memory
capacity of 512, the qualifier-aware StarE-GNN configuration achieves the best held-out
performance among the compared symbolic, neural, and neuro-symbolic systems while
preserving step-level traceability of memory-management decisions.

\end{abstract}

\section{Introduction}
Reinforcement learning (RL) in partially observable environments requires effective memory
management: the agent must decide what to retain, what to forget, and how to retrieve
task-relevant information over long horizons. Classical deep RL approaches improve
control performance with function approximation
\cite{mnih2013playingatarideepreinforcement,Vincent_2018}, and recurrent extensions
mitigate partial observability by integrating observation histories
\cite{hausknecht2017deeprecurrentqlearningpartially,10.1162/neco.1997.9.8.1735}.
However, latent memory in recurrent or end-to-end neural systems is often difficult to
inspect and trace, which limits their suitability when explicit reasoning and
explanation are important.

At the same time, symbolic and knowledge-graph-based approaches show that structured
representations can improve transfer and make control decisions easier to inspect
\cite{garcez2018symbolicreinforcementlearningcommon,zambaldi2018relationaldeepreinforcementlearning}.
For graph-structured representations, relation-aware encoders such as the relational
graph convolutional network (R-GCN) and qualifier-aware hyper-relational message passing
are especially relevant
\cite{schlichtkrull2017modelingrelationaldatagraph,galkin2020messagepassinghyperrelationalknowledge}.
In our setting, hidden state and observations are represented as Resource Description
Framework (RDF) graphs \cite{cyganiak2014rdf11concepts,hayes2014rdf11semantics,rdf12concepts,rdf12turtle}. Memory
additionally uses temporal annotation properties on observed facts, enabling explicit
time- and provenance-like metadata over the memory graph. Concretely, the agent must
answer object-location queries while navigating from partial room-level observations
under a bounded long-term memory budget, so the core problem is how to coordinate
retrieval, exploration, and forgetting when not all relevant facts can be retained.
This provides the paper's Semantic Web grounding: the agent's internal state is an
RDF-based knowledge graph with explicit annotations, and its operators act over that
graph through retrieval and update procedures.

This structure creates a practical design tension. Fixed symbolic heuristics (e.g.,
predetermined policies for forgetting or retrieval) are explicit but inflexible across
contexts. Fully neural policies are adaptive but typically opaque.

We build on the RoomKG temporal knowledge-graph memory substrate. The environment, its
episodes, its symbolic memory heuristics, and the temporally annotated RDF memory
representation are prior work \cite{kim2026temporalknowledgegraphmemorypartially}. On
top of this substrate, we treat memory decisions as a meta-decision problem: a learned
meta-policy selects among named symbolic heuristics at each decision point. The
contribution of this paper is this meta-policy layer and its controlled evaluation, not
the benchmark or the memory substrate itself. This perspective is related to hierarchical policy
selection and options \cite{SUTTON1999181,bacon2016optioncriticarchitecture}, but here
the selectable actions are explicit symbolic operators rather than latent option
embeddings. In the bounded-memory RoomKG setting we study, this lets us ask a narrower
question than broad end-to-end superiority: can learned heuristic selection improve
adaptivity over strong fixed symbolic controls while preserving explicit symbolic
execution, and are qualifier-aware graph encoders especially well matched to that
decision problem?

Our main contributions are as follows:
\begin{itemize}
	\item We introduce a neuro-symbolic \emph{meta-policy} formulation for
	partially observable RL in which adaptation is achieved by selecting among
	named symbolic memory heuristics, not by replacing symbolic control with
	opaque latent actions.
	\item We design a qualifier-aware temporal knowledge-graph memory
	meta-policy that combines an RDF-based memory representation with RDF triple
	annotations, knowledge-graph-based encoding, and task-specific value
	heads for question answering, exploration, and forgetting.
	\item We demonstrate strong train/test generalization in RoomKG at long-term
	memory capacity of 512, with the StarE-GNN-based configuration delivering the best
	overall performance while preserving step-level traceability of
	memory-management decisions.
\end{itemize}

\section{Background and Problem Setting}

\begin{figure}[tb]
\centering
\includegraphics[width=0.8\linewidth]{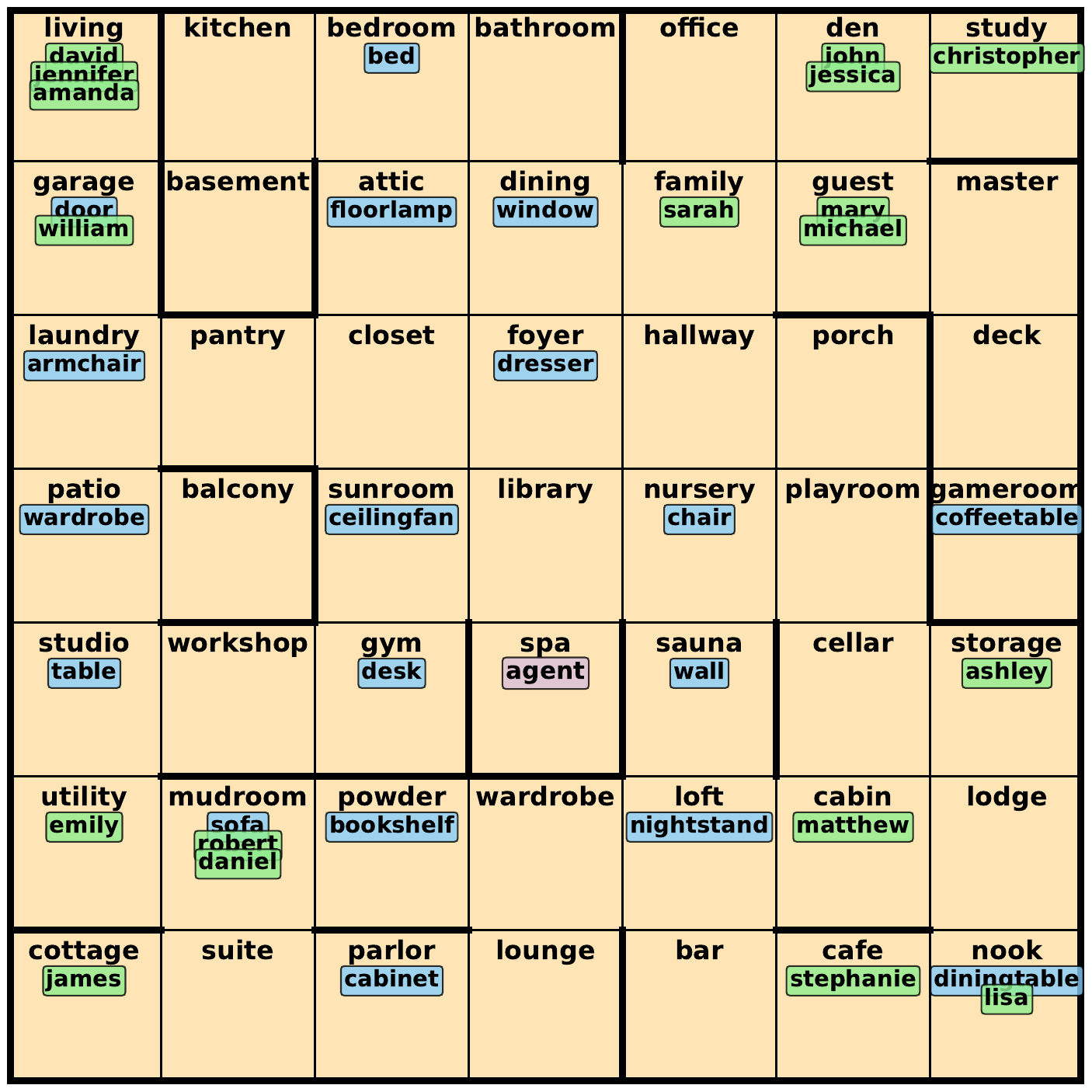}
\caption{Bird's-eye schematic of the hidden state at $t=99$ ($s_{t=99}$) in RoomKG,
showing spatial layout and entity placement. This view is only a schematic for
intuition: the actual environment state and agent-facing world are represented in the
RDF knowledge-graph world (Figure~\ref{fig:hidden-state-rdf-step99}). The agent does not
directly observe $s_t$; its observation $o_t$ is the induced RDF subgraph of the current
room and visible adjacency relations. The figure helps ground the later graph view in an
intuitive spatial picture, making clear why partial observability is nontrivial even in
a structured environment. The full original figure is provided in the supplemental
material.}
\label{fig:bird-eye-view-step99}
\end{figure}

\begin{figure}[tb]
\centering
\includegraphics[width=0.88\linewidth]{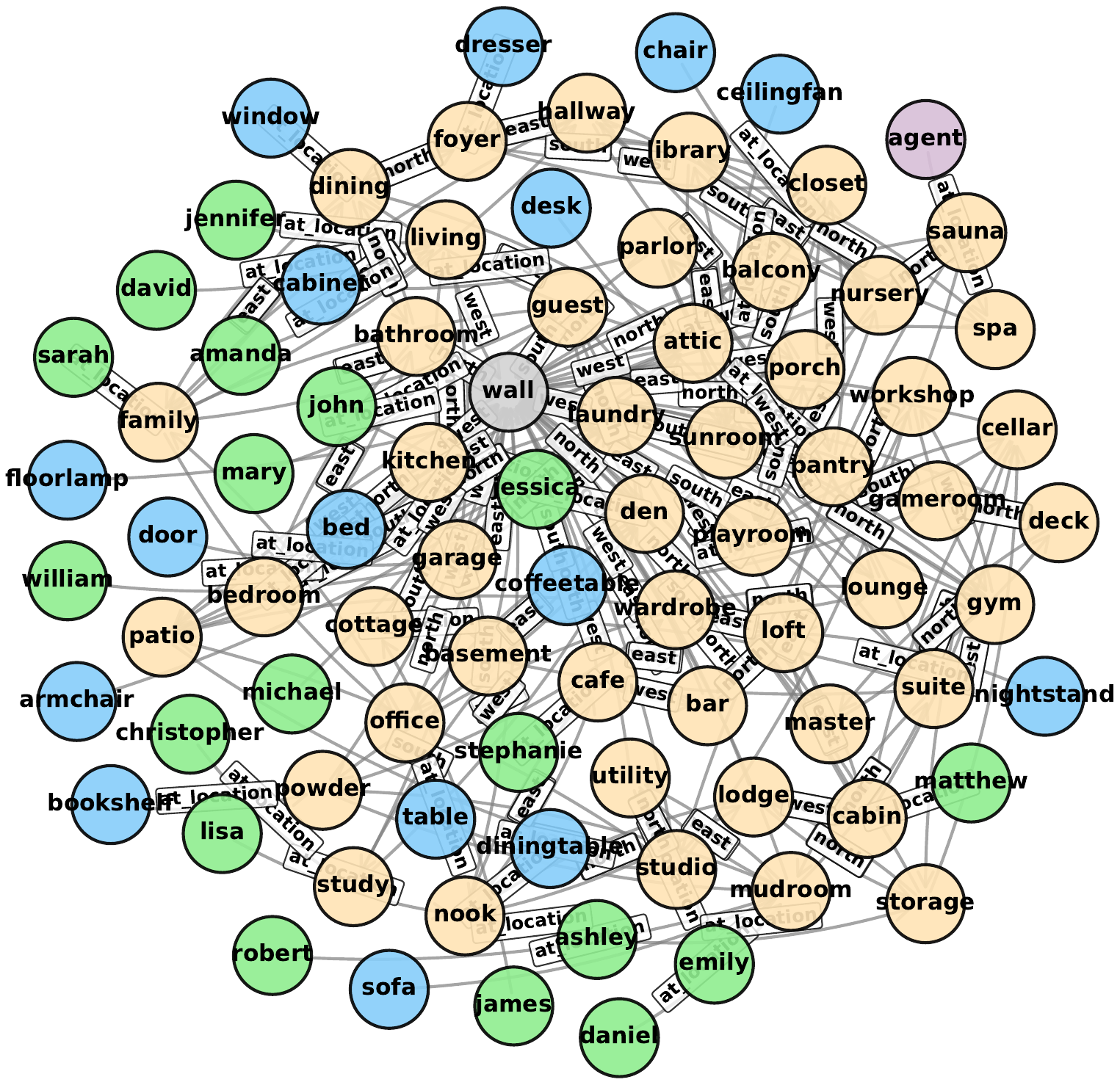}
\caption{Knowledge-graph view of the same hidden state at $t=99$ ($s_{t=99}$), expressed
as RDF structure. Node colors indicate semantic categories: rooms (yellow), agent (purple),
static objects (blue), moving objects (green), and walls (grey). This is the structural form
that motivates relation-aware encoders and explicit memory annotations in the proposed
meta-policy. The full original figure is provided in the supplemental material.}
\label{fig:hidden-state-rdf-step99}
\end{figure}

\subsection{Environment and Partial Observability}
We consider the RoomKG benchmark \cite{kim2026temporalknowledgegraphmemorypartially}, a sequential
decision-making environment in which the agent receives only partial observations while
the full environment state is hidden. Let $s_t$ denote the hidden state at time $t$ and
$o_t$ the raw observation. The agent must act using a history-dependent internal memory,
because $o_t$ alone is generally insufficient to infer all task-relevant facts. This
places the problem in the standard partially observable setting, where performance
depends critically on what information is stored, updated, and retrieved over time.

Concretely, RoomKG is a grid world whose hidden state is an RDF knowledge graph over
rooms, objects, walls, and the agent. At each step, the observation $o_t$ is the RDF
subgraph induced by the current room (local adjacency and object-location triples). Each
step also issues a query asking for the current location of a named object, and the
environment-facing action consists of an answer plus a movement choice from \{north,
east, south, west, stay\}. This design forces agents to integrate partial observations
over time to answer correctly. In an end-to-end neural formulation, this becomes a joint
discrete action problem over 49 candidate room answers and 5 movement directions (245
actions).

The hidden dynamics are structured but nontrivial: inner-wall availability follows fixed
periodic schedules, and moving objects follow deterministic preference rules subject to
wall constraints. This keeps the benchmark replayable while still requiring long-horizon
memory under local observability.

Formally, we model interaction as a partially observable process
\cite{kaelbling1998planning}
\[
\langle \mathcal{S}, \mathcal{O}, \mathcal{A}, P, R, \Omega, \gamma \rangle,
\]
where $\mathcal{S}$ is the hidden-state space, $\mathcal{O}$ the observation space,
$\mathcal{A}$ the action space, $P(s_{t+1}\mid s_t,a_t)$ the transition model,
$R(s_t,a_t)$ the reward function, $\Omega(o_t\mid s_t)$ the observation model, and
$\gamma\in[0,1)$ the discount factor. The agent does not have direct access to $s_t$ at
decision time, and therefore must approximate belief-relevant information through an
explicit memory representation. In RoomKG, this requirement is particularly
important because useful cues can be temporally delayed and distributed across distinct
observations.

Our focus is the memory-decision layer that determines which symbolic operation to apply
given the current observation and memory contents.

\subsection{Temporal Knowledge-Graph Memory Representation}

Both hidden states and observations are represented as RDF graphs grounded in RDF
\cite{cyganiak2014rdf11concepts,hayes2014rdf11semantics,rdf12concepts,rdf12turtle}. The agent maintains a temporal
knowledge-graph memory $M_t$ that extends observed facts with temporal annotation
properties using RDF triple annotations: \texttt{time\_added},
\texttt{last\_accessed}, and \texttt{num\_recalled}. This makes memory contents explicit
and inspectable at the level of annotated triples. In other words, each remembered fact
is stored as a base RDF triple together with statement-level metadata about when it was
added, last used, and how often it has been recalled. We use this annotation
mechanism as the implementation substrate for explicit memory metadata, not as a new
contribution to temporal RDF semantics.

For example, an observation such as \texttt{(agent, at\_location, room)} can enter
short-term memory with a \texttt{current\_time} annotation at step $t$ and, if retained,
the same RDF triple carries annotations such as \texttt{time\_added}$=t$,
\texttt{last\_accessed}$=t'$, and \texttt{num\_recalled}$=n$, so the agent's memory
remains an explicit RDF graph with statement-level temporal metadata.

We keep separate short- and long-term memories, $M_t^{\mathrm{short}}$ and
$M_t^{\mathrm{long}}$, with $M_t=(M_t^{\mathrm{short}}, M_t^{\mathrm{long}})$. Raw
observations $o_t$ become short-term items by attaching \texttt{current\_time}, and
symbolic remember/forget heuristics either refresh long-term entries with these items or
evict long-term items when capacity is hit.

At each step, observed RDF triples are merged into memory according to the selected
symbolic heuristic. Qualifiers encode insertion time, access recency, and recall
frequency, so the agent distinguishes facts both by graph content and by temporal role.
Long-term memory is bounded, $|M_t^{\mathrm{long}}|\leq K$, so control must coordinate
question answering (QA), exploration, and forgetting under explicit capacity pressure.

Figures~\ref{fig:bird-eye-view-step99} and \ref{fig:hidden-state-rdf-step99} ground the
representational gap the controller must bridge: the hidden world has rich relational
structure, while the agent receives only room-local RDF observations and must maintain
its own bounded memory over time.

\subsection{Symbolic Heuristic Action Space}

Instead of outputting opaque latent actions, the meta-policy selects among named
symbolic heuristics for QA, exploration, and forgetting.

Let $\mathcal{H}^{\mathrm{forget}}$, $\mathcal{H}^{\mathrm{qa}}$, and
$\mathcal{H}^{\mathrm{explore}}$ denote the candidate sets for the three decision
categories. At each decision point, the meta-policy selects one heuristic from each set,
yielding a joint symbolic action $h_t=(h_t^{\mathrm{forget}}, h_t^{\mathrm{qa}},
h_t^{\mathrm{explore}})$ in the finite product space
$\mathcal{H}^{\mathrm{forget}}\times\mathcal{H}^{\mathrm{qa}}\times
\mathcal{H}^{\mathrm{explore}}$.

Rather than learning the full environment-facing combinatorial action directly, the
meta-policy operates over a modular $3\times3\times3=27$ symbolic space. The operators
themselves are fixed RoomKG primitives: QA retrieves over annotated RDF memory,
exploration plans over the remembered map, and forgetting applies cache-like eviction
rules. The learned component is which heuristic to activate online.

Operationally, QA uses most-recently-added (MRA), most-recently-used (MRU), and
most-frequently-used (MFU) ranking over qualifier metadata, exploration uses the same
qualifier priorities to filter remembered graph structure before planning, and
forgetting uses first-in-first-out (FIFO), least-recently-used (LRU), and
least-frequently-used (LFU) eviction when capacity is reached.

\subsection{Problem Formulation}

At each decision point, a parametric meta-policy $\pi_\theta(h_t \mid M_t)$ maps memory
state $M_t$ to a joint symbolic action
$h_t=(h_t^{\mathrm{forget}}, h_t^{\mathrm{qa}}, h_t^{\mathrm{explore}})$. In the
partially observable Markov decision process (POMDP) setting, $M_t$ is a bounded
internal summary of observation history rather than an exact belief state.

We write the symbolic execution step abstractly as
\begin{equation}
M_t^{\mathrm{short}} = \mathrm{Aug}(o_t),\qquad
M_t = (M_t^{\mathrm{short}}, M_t^{\mathrm{long}}),
\label{eq:aug-memory}
\end{equation}
followed by
\begin{equation}
(M_{t+1}, a_t) = \Phi\big(M_t, h_t\big),
\label{eq:phi-exec}
\end{equation}
where $\mathrm{Aug}(\cdot)$ converts raw observation triples into short-term memory items
by attaching \texttt{current\_time}, and $\Phi$ deterministically executes the selected
heuristics to update memory and choose environment-facing behavior.

Training maximizes expected return under a bounded-memory constraint,
\begin{equation}
\max_{\theta}\; \mathrm{E}_{\pi_\theta}\!\left[\sum_{t=0}^{T} \gamma^t r_t\right]
\quad \text{s.t.} \quad |M_t^{\mathrm{long}}| \leq K \; \forall t,
\end{equation}

Here, $\gamma$ is the discount factor and $r_t$ is the reward at time $t$; in RoomKG,
reward is generated only by question-answering correctness ($+1$ for correct answers,
$0$ otherwise), so exploration and forgetting are learned through their downstream
contribution to future QA outcomes.

\section{Method}

\subsection{Overview}

Our meta-policy takes memory state $M_t$ and predicts symbolic heuristic choices for
QA, exploration, and forgetting. Neural components score candidate symbolic heuristics;
deterministic symbolic procedures execute the selected ones. Concretely, the pipeline is
to encode memory contents, score heuristic candidates, select heuristics with
$\epsilon$-greedy exploration during training, and execute them through $\Phi$
(Algorithm~\ref{alg:control-loop}).

\begin{algorithm}[t]
\caption{One-step neuro-symbolic decision loop}
\label{alg:control-loop}
\begin{algorithmic}[1]
\Require long-term memory $M_t^{\mathrm{long}}$, observation $o_t$, question $q_t$,
heuristic sets $\mathcal{H}^{\mathrm{qa}},\mathcal{H}^{\mathrm{explore}},\mathcal{H}^{\mathrm{forget}}$,
parameters $\theta$, exploration rate $\epsilon$ ($\epsilon{=}0$ at test time)
\Ensure updated memory $M_{t+1}$, environment action $a_t$
\State $M_t^{\mathrm{short}} \gets \mathrm{Aug}(o_t)$;\quad
$M_t \gets (M_t^{\mathrm{short}}, M_t^{\mathrm{long}})$
\Comment{Eq.~\eqref{eq:aug-memory}}
\State $\mathbf{Z}_t \gets \mathrm{Enc}_\theta(M_t)$
\Comment{GCN / R-GCN / StarE-GNN}
\For{$p \in \{\mathrm{qa}, e, f\}$}
\State $\mathbf{m}^{(p)}_t \gets \sum_j \alpha^{(p)}_{t,j} \mathbf{W}_V \mathbf{z}_{t,j}$
\Comment{head-specific attention pooling}
\EndFor
\State $h_t^{\mathrm{qa}} \gets
\epsilon\text{-}\mathrm{greedy}_{h\in\mathcal{H}^{\mathrm{qa}}}\,
Q_{\mathrm{qa}}([\mathbf{m}^{(qa)}_t,\phi(q_t)],h)$
\State $h_t^{\mathrm{explore}} \gets
\epsilon\text{-}\mathrm{greedy}_{h\in\mathcal{H}^{\mathrm{explore}}}\,
Q_e(\mathbf{m}^{(e)}_t,h)$
\State $h_t^{\mathrm{forget}} \gets
\epsilon\text{-}\mathrm{greedy}_{h\in\mathcal{H}^{\mathrm{forget}}}\,
Q_f(\mathbf{m}^{(f)}_t,h)$
\State $h_t \gets (h_t^{\mathrm{forget}}, h_t^{\mathrm{qa}}, h_t^{\mathrm{explore}})$
\Comment{or one 27-way head}
\State $(M_{t+1}, a_t) \gets \Phi(M_t, h_t)$
\Comment{symbolic execution, Eq.~\eqref{eq:phi-exec}}
\If{training}
\State observe $r_t$; append transition to $\mathcal{D}_p$;
update $\theta_p$ by $\mathcal{L}^{(p)}$
\Comment{Eq.~\eqref{eq:head-loss}}
\EndIf
\end{algorithmic}
\end{algorithm}

\paragraph{Worked example.}
Suppose at $t{=}12$ long-term memory holds
$f_1$ = \texttt{(sarah, at\_location, living)} with
(\texttt{time\_added}, \texttt{last\_accessed}, \texttt{num\_recalled}) = $(4,11,5)$,
$f_2$ = \texttt{(sarah, at\_location, kitchen)} with $(10,10,1)$, and
$f_3$ = \texttt{(floorlamp, at\_location, attic)} with $(2,9,3)$.
For the query ``where is \texttt{sarah}?'', MRA answers \texttt{kitchen} (largest
\texttt{time\_added} among matching facts), whereas MRU and MFU answer \texttt{living};
if Sarah recently moved, only MRA is correct. If a new observation arrives at
capacity $K{=}3$, LFU evicts $f_2$ (fewest recalls), while LRU and FIFO evict $f_3$
(oldest access; earliest insertion). The learned heads score exactly these named
alternatives, so every decision remains attributable to an explicit heuristic over
explicit annotations.

\subsection{Knowledge-Graph Memory Encoding}

Each memory item is represented as a knowledge-graph fact with qualifiers (e.g.,
\texttt{current\_time}, \texttt{time\_added}, \texttt{last\_accessed}, \texttt{num\_recalled}). We encode the
working memory graph with one of three interchangeable function approximators: a graph
convolutional network (GCN), an R-GCN, or a StarE graph neural network (StarE-GNN)
\cite{kipf2017semisupervisedclassificationgraphconvolutional,schlichtkrull2017modelingrelationaldatagraph,galkin2020messagepassinghyperrelationalknowledge,zambaldi2018relationaldeepreinforcementlearning}.
We compare them to isolate the contribution of topology-only, relation-aware, and
qualifier-aware encoding, respectively.
For graph encoders, we include inverse relations and qualifier relations in the relation
vocabulary.

Because memory is a variable-size knowledge graph, we use message-passing GNN variants
(GCN, R-GCN, StarE-GNN) that operate naturally on non-fixed graph structure
\cite{schlichtkrull2017modelingrelationaldatagraph,galkin2020messagepassinghyperrelationalknowledge,zambaldi2018relationaldeepreinforcementlearning}. The encoder outputs contextual embeddings reused by policy-specific heads.

For GNN encoders, we compute node embeddings by message passing on the memory graph.
GCN is the simplest case and ignores relation types and qualifiers, using only
topology:
\begin{equation}
\mathbf{g}^{(k)}_v = \sigma\!\left(\sum_{u\in\mathcal{N}(v)} \mathbf{W}^{(k)}\mathbf{g}^{(k-1)}_u\right).
\end{equation}
R-GCN conditions on relation types but does not encode qualifier values:
\begin{equation}
\mathbf{g}^{(k)}_v = \sigma\!\left(\sum_{r\in\mathcal{R}}\sum_{u\in\mathcal{N}_r(v)}
\frac{1}{c_{v,r}} \mathbf{W}^{(k)}_r\mathbf{g}^{(k-1)}_u\right).
\end{equation}
StarE-GNN extends R-GCN by incorporating qualifier embeddings $\mathbf{q}_{u,r,v}$ in the
message function, enabling qualifier-aware aggregation:
\begin{equation}
\mathbf{g}^{(k)}_v = \sigma\!\left(\sum_{(u,r)\in\mathcal{N}(v)}
\mathbf{W}^{(k)}_{\lambda(r)}\,\phi_r\!\big(\mathbf{g}^{(k-1)}_u,\,\psi(\mathbf{g}_r,\mathbf{q}_{u,r,v})\big)\right).
\end{equation}
Here, $\mathcal{N}(v)$ and $\mathcal{N}_r(v)$ denote the (relation-agnostic and
relation-specific) neighbor sets of node $v$, $\sigma$ is a pointwise nonlinearity,
$\mathbf{W}^{(k)}$ and $\mathbf{W}_r^{(k)}$ are layer-$k$ trainable weights, $c_{v,r}$ is the R-GCN
normalization constant, $\mathbf{g}_r$ is the embedding of relation $r$,
$\mathcal{R}$ is the set of relation types, and $k$ indexes message-passing layers,
$\lambda(r)$ maps directed edges to relation types (including inverse types), and
$\phi_r(\cdot)$ and $\psi(\cdot)$ denote relation-specific message composition and
qualifier-combination functions, respectively.
The encoder maps the memory to contextual vectors
\begin{equation}
\mathbf{Z}_t=\mathrm{Enc}_\theta(M_t)=\{\mathbf{z}_{t,j}\}_{j=1}^{|M_t|}.
\end{equation}
To feed fixed-size value heads, we apply head-specific attention pooling
\cite{bahdanau2016neuralmachinetranslationjointly}
\begin{align}
\alpha^{(p)}_{t,j}
&=\frac{\exp\!\left((\mathbf{q}^{(p)})^\top \mathbf{W}_K\mathbf{z}_{t,j}/\sqrt{d}\right)}
{\sum_{\ell}\exp\!\left((\mathbf{q}^{(p)})^\top \mathbf{W}_K\mathbf{z}_{t,\ell}/\sqrt{d}\right)},\\
\mathbf{m}^{(p)}_t
&=\sum_j \alpha^{(p)}_{t,j}\mathbf{W}_V\mathbf{z}_{t,j},
\end{align}
for $p\in\{\mathrm{qa},e,f\}$ (QA, explore, forget).
Here $\alpha^{(p)}_{t,j}$ is the attention weight that head $p$ assigns to memory item
$j$ at time $t$, and $\mathbf{m}^{(p)}_t$ is the resulting head-specific pooled memory
summary that serves as the fixed-size input to head $p$'s value function;
$\mathbf{q}^{(p)}$ is a learned query vector for head $p$, $\mathbf{W}_K$ and $\mathbf{W}_V$ are
learned key/value projections, and $d$ is the embedding dimension.

\subsection{Heuristic-Selection Heads}

Let $\mathbf{m}^{(qa)}_t$, $\mathbf{m}^{(e)}_t$, and $\mathbf{m}^{(f)}_t$ denote the
head-specific pooled contexts above. We define three value functions over symbolic
heuristic sets:
\begin{align}
Q_{\mathrm{qa}}([\mathbf{m}^{(qa)}_t,\phi(q_t)], h^{\mathrm{qa}}),\quad
Q_e(\mathbf{m}^{(e)}_t, h^{\mathrm{explore}}),\quad
Q_f(\mathbf{m}^{(f)}_t, h^{\mathrm{forget}}),
\end{align}
where $Q_{\mathrm{qa}}$, $Q_e$, and $Q_f$ are the QA-, exploration-, and forget-head action-value
functions, respectively. Here, $q_t$ is the current question tuple and $\phi(\cdot)$ is
its learned embedding used to condition the QA head. In the independent-head
configuration, these heads output values over $|\mathcal{H}^{\mathrm{qa}}|=3$
(most-recently-added (MRA), most-recently-used (MRU), most-frequently-used (MFU)),
$|\mathcal{H}^{\mathrm{explore}}|=3$ (MRA/MRU/MFU), and $|\mathcal{H}^{\mathrm{forget}}|=3$
(first-in-first-out (FIFO), least-recently-used (LRU), least-frequently-used (LFU)).

The corresponding greedy selections are
\begin{align}
h_t^{\mathrm{qa}}
&= \operatorname*{arg\,max}_{h\in\mathcal{H}^{\mathrm{qa}}}
Q_{\mathrm{qa}}([\mathbf{m}^{(qa)}_t,\phi(q_t)],h),\\
h_t^{\mathrm{explore}}
&= \operatorname*{arg\,max}_{h\in\mathcal{H}^{\mathrm{explore}}}
Q_e(\mathbf{m}^{(e)}_t,h),\\
h_t^{\mathrm{forget}}
&= \operatorname*{arg\,max}_{h\in\mathcal{H}^{\mathrm{forget}}}
Q_f(\mathbf{m}^{(f)}_t,h),
\end{align}
with $\epsilon$-greedy exploration during training for each head.

At decision time, we select
\begin{equation}
h_t = \big(h_t^{\mathrm{forget}}, h_t^{\mathrm{qa}}, h_t^{\mathrm{explore}}\big)
\end{equation}
by $\epsilon$-greedy maximization of the corresponding head outputs. The implementation
also supports a combinatorial variant with a single 27-way head over
$\mathcal{H}^{\mathrm{forget}}\times\mathcal{H}^{\mathrm{qa}}\times\mathcal{H}^{\mathrm{explore}}$.

\subsection{Symbolic Execution and Learning}
\label{sec:execution-learning}

Given $h_t$, symbolic execution applies the selected heuristics using the same
observation-ingestion and transition definitions in
Eqs.~\eqref{eq:aug-memory}--\eqref{eq:phi-exec}. In our implementation, short-term
ingestion is deterministic in the main setting, while learned control focuses on
QA, exploration, and forgetting. Here, $\Phi$ is simply shorthand for this fixed
symbolic execution routine.

Training uses off-policy replay with a target network for temporal-difference updates
with $\epsilon$-greedy exploration and periodic target-network synchronization. Forget
and explore heads are trained with bootstrapped temporal-difference updates, while QA is
trained as an immediate-reward (contextual-bandit) head.

The overall training objective is the sum of active per-head losses. For each head
$p\in\{\mathrm{qa},e,f\}$, we use a unified deep Q-network (DQN)-style objective
\cite{mnih2013playingatarideepreinforcement}
\begin{equation}
\label{eq:head-loss}
\mathcal{L}^{(p)}(\theta_p)=
\mathrm{E}_{(x,a,r,x',d)\sim \mathcal{D}_p}
\left[
\left(y^{(p)}-Q^{(p)}_{\theta_p}(x,a)\right)^2
\right],
\end{equation}
with target
\begin{equation}
y^{(p)} = r + \gamma_p(1-d)\max_{a'}Q^{(p)}_{\theta_p^-}(x',a').
\end{equation}
Here $\mathcal{D}_p$ is the replay distribution for head $p$, $\theta_p$ and
$\theta_p^-$ are online and target parameters, and $d\in\{0,1\}$ is the terminal
indicator. We set $\gamma_{\mathrm{qa}}=0$ for QA, so the QA target reduces to immediate
reward, while $\gamma_f,\gamma_e>0$ retain temporal credit assignment for forget and
explore.

For notation, a training sample for forget/explore is $(x,a,r,x',d)$ with memory-state
input $x$, symbolic heuristic action $a$, reward $r$, next input $x'$, and terminal
indicator $d\in\{0,1\}$; for QA, the head input is question-conditioned (memory plus
question embedding).

We model QA as contextual bandit because its decision is primarily
tied to immediate answer correctness and has only limited downstream state effect
(mainly qualifier updates such as access/recall metadata). In contrast, exploration and
forgetting alter which memory items are retained or encountered, so their consequences
are genuinely long-horizon and require TD (temporal difference) credit assignment. Because RoomKG reward
is emitted only by QA correctness, exploration and forgetting heuristics are learned
through delayed credit assignment via their downstream effect on future QA reward.

\section{Experimental Setup}
\label{sec:experimental-setup}
We evaluate in RoomKG using
the standard \texttt{train}/\texttt{test} split: \texttt{train} for
training/model selection and \texttt{test} as the held-out environment (same
underlying dynamics with permuted question order). Episodes terminate at step 99
(100 interaction steps).

\subsection{Compared Methods}

We compare three families.
\begin{itemize}
	\item \textbf{Symbolic-TKG}: temporal annotated RDF memory with fixed symbolic
	heuristics for QA/\allowbreak exploration/\allowbreak forgetting.
	\item \textbf{Neural (Simple DQN)}: end-to-end sequence agent (long
	short-term memory (LSTM) or Transformer)
	\cite{10.1162/neco.1997.9.8.1735,vaswani2023attentionneed}
	that directly predicts the environment-level combinatorial action
	(49 rooms $\times$ 5 directions = 245 discrete actions).
	\item \textbf{Neuro-Symbolic TKG}: shared symbolic memory/execution with
	learned heuristic selection heads, using one of
	\{GCN, R-GCN, StarE-GNN\} as the encoder backbone.
\end{itemize}

Neuro-Symbolic TKG is our contribution; the other families are in-domain reference
systems from prior RoomKG work. For the neuro-symbolic model, we evaluate both
independent policy heads and a 27-way combinatorial head, and both shared-network and
separate-network training variants. We also include stronger non-learning symbolic
controls: memory pressure (MP), random meta (RM), intersection vote (IV), and ranked
voting (RankVote). These serve as additional reference policies for testing learned
meta-selection.

\begin{table}[tb]
\centering
\small
\setlength{\tabcolsep}{3pt}
\resizebox{\linewidth}{!}{%
\begin{tabular}{llllllll}
\toprule
\textbf{Method} & \textbf{Memory} & \textbf{Encoder} & \textbf{QA} & \textbf{Explore} & \textbf{Forget} & \textbf{Train QA} & \textbf{Test QA} \\
\midrule
S-TKG & Ann. RDF & -- & MRU & MRU & LRU & 44.48 (1.23) & 45.64 (1.21) \\
S-TKG & Ann. RDF & -- & MP & MP & MP & 43.60 (1.11) & 45.88 (2.32) \\
S-TKG & Ann. RDF & -- & RM & RM & RM & 43.36 (0.83) & 46.28 (1.46) \\
S-TKG & Ann. RDF & -- & RV & RV & RV & 43.32 (1.12) & 44.16 (1.52) \\
S-TKG & Ann. RDF & -- & IV & IV & IV & 43.08 (1.40) & 46.28 (0.89) \\
Neural (E2E) & ObsHist & LSTM & RL & RL & FIFO & 13.20 (2.04) & 9.40 (2.58) \\
Neural (E2E) & ObsHist & Transf. & RL & RL & FIFO & 18.60 (0.49) & 11.40 (2.42) \\
NS-TKG & Ann. RDF & GCN & RL & RL & RL & 46.92 (0.41) & 46.04 (1.44) \\
NS-TKG & Ann. RDF & R-GCN & RL & RL & RL & 47.08 (0.30) & 47.00 (1.67) \\
NS-TKG & Ann. RDF & StarE-GNN & RL & RL & RL & 46.72 (0.89) & \textbf{47.28 (0.98)} \\
NS-TKG & Ann. RDF & StarE-GNN$^{s}$ & RL & RL & RL & \textbf{47.16 (0.66)} & 47.16 (1.49) \\
NS-TKG & Ann. RDF & StarE-GNN$^{c}$ & RL & RL & RL & 42.28 (1.25) & 46.12 (2.24) \\
\bottomrule
\end{tabular}
}
\caption{Main train/test results at long-term memory capacity of 512 (mean with
standard deviation in parentheses). Abbreviations: symbolic temporal knowledge graph
(S-TKG), neuro-symbolic temporal knowledge graph (NS-TKG), annotated RDF (Ann. RDF),
end-to-end (E2E), observation history (ObsHist), memory pressure (MP), random meta (RM),
intersection vote (IV), ranked voting (RV), and learned independent heads (RL).
For the E2E neural baselines, RL denotes the joint DQN policy over composite actions.
BFS denotes breadth-first exploration in the symbolic baselines. StarE-GNN$^{s}$ is the
shared-network ablation, and StarE-GNN$^{c}$ is the learned combinatorial-head ablation.}
\label{tab:main-results}
\end{table}

\subsection{Training Protocol}

All learned agents use off-policy Q-learning with replay and target-network updates
\cite{mnih2013playingatarideepreinforcement}. For the main experiments, each run uses
200 training episodes (20{,}000 environment steps), batch size 32, replay buffer size
equal to the training horizon, warm-start of 10\% of training steps, target update
interval 50, and linear $\epsilon$-decay from 1.0 to 0.01 over the first half of
training. We use learning rate $10^{-4}$, discount $\gamma=0.95$ (with
head-specific handling for QA as described in Section~\ref{sec:execution-learning}), and gradient clipping.

\subsection{Memory-Capacity Settings and Reporting Scope}

Kim et al. \cite{kim2026temporalknowledgegraphmemorypartially} report the
long-term memory-capacity sweep from 0 to 512. Following that setup, we
focus this paper on 512, which is large enough to expose meaningful differences
between heuristic-selection strategies and encoder choices while still enforcing a
strict capacity bound.

\subsection{Evaluation Metrics}

Episode return corresponds to cumulative QA success under the reward defined in the
problem formulation. We report test performance on the held-out environment as mean and
standard deviation across 5 random seeds; validation follows the same train-environment
protocol for model selection. Supporting reproducibility details are provided in the
supplemental material package.

\section{Results}

\begin{figure}[tb]
\centering
\includegraphics[width=0.6\linewidth]{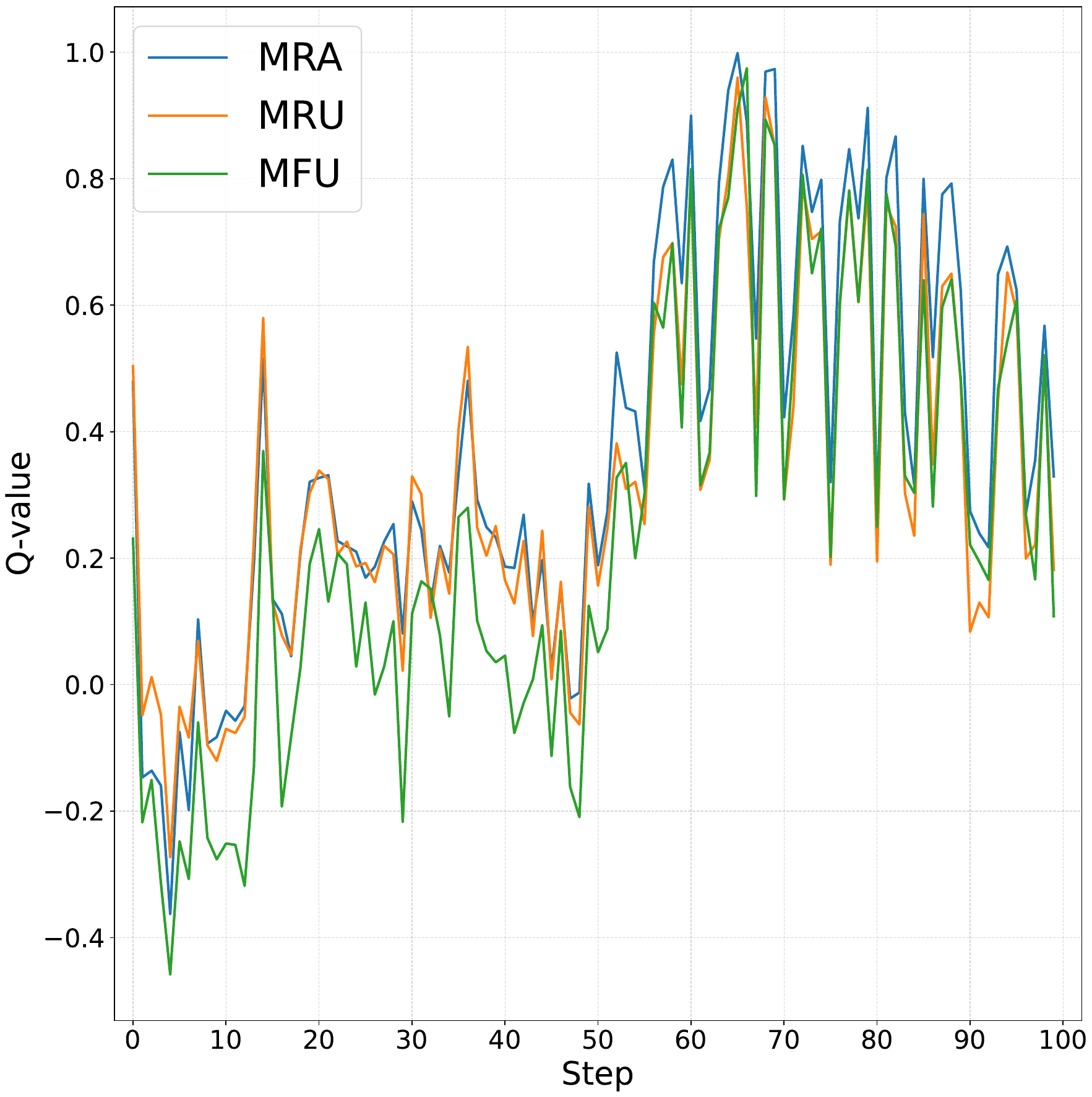}
\caption{Per-step Q-value trajectories for the QA head of the learned StarE-GNN
neuro-symbolic model on the held-out \texttt{test} trace. Each curve corresponds to one
symbolic QA heuristic, showing how preference shifts over semantically named choices
rather than over opaque latent actions. The full three-head figure set, including
exploration and forgetting, is provided in the supplemental material.}
\label{fig:q-values-qa}
\end{figure}

This section is organized as three nested comparisons. Section~\ref{sec:main-results}
compares the three method families (symbolic, neural, neuro-symbolic) in
Table~\ref{tab:main-results}. Section~\ref{sec:ablations} ablates within the
neuro-symbolic family along two axes: encoder expressiveness (topology-only GCN vs.
relation-aware R-GCN vs. qualifier-aware StarE-GNN) and head structure (modular
three-head vs. shared-network StarE-GNN$^{s}$ vs. combinatorial 27-way
StarE-GNN$^{c}$). Section~\ref{sec:symbolic-diagnostics} examines the spread across the
fixed symbolic controls.

\subsection{Main Quantitative Results}
\label{sec:main-results}

Table~\ref{tab:main-results} shows three clear patterns. First, the qualifier-aware
neuro-symbolic variants are strongest overall on held-out test, with StarE-GNN best and
R-GCN close behind. Second, the end-to-end neural baselines are far weaker than both
symbolic-TKG and neuro-symbolic TKG, indicating that direct 245-way action prediction is
poorly matched to this bounded-memory setting. Third, the best symbolic controls remain
strong, but learned neuro-symbolic selection still gives the best held-out result.

Because reward is defined through QA correctness, train/test QA in
Table~\ref{tab:main-results} is the central end-to-end metric: improvements reflect
better coordination between forgetting, exploration, and retrieval. The strongest gains
also align with encoder design. StarE-GNN and R-GCN preserve the relation and qualifier
signals that the symbolic heuristics depend on, whereas topology-only GCN and the direct
combinatorial ablation are less well aligned with the decision problem.

\subsection{Ablations: Neuro-Symbolic Variants}
\label{sec:ablations}

Within the neuro-symbolic family, the main pattern is that relation-aware and
qualifier-aware encoders outperform topology-only aggregation, with R-GCN and
StarE-GNN strongest on held-out test. The shared-network StarE-GNN$^{s}$ is close to the
main StarE-GNN configuration, so the shared-vs-separate difference is small in absolute
QA terms. By contrast, the combinatorial StarE-GNN$^{c}$ is weaker and more variable,
supporting the modular three-head design over direct 27-way symbolic control.

This ablation pattern is consistent with the task structure: the heuristics depend on
relation and qualifier semantics, so encoders that preserve those signals are better
matched to the decision problem than topology-only aggregation. It is also consistent
with the optimization setting: three independent 3-way heads receive denser supervision
than one joint 27-way action head under the same reward stream.

\subsection{Symbolic Baseline Diagnostics}
\label{sec:symbolic-diagnostics}

The fixed symbolic baselines in Table~\ref{tab:main-results} remain competitive, which
shows that the underlying symbolic memory substrate is already strong. At the same time,
their spread in test performance shows that heuristic choice matters even without
learning. This supports the paper's core premise: in partially observable tasks,
\emph{which} symbolic heuristic is active is itself a consequential decision variable.
Against this stronger non-learning baseline set, the best learned neuro-symbolic model
still remains highest on held-out test, supporting learned online selection over fixed
global combination rules.

\section{Qualitative Analysis and Decision Traces}

\subsection{QA-Value Dynamics}

Figure~\ref{fig:q-values-qa} shows that QA heuristic preferences evolve over
the episode in an interpretable way. The QA head remains mostly in the $[0,1]$ range,
consistent with its immediate-reward formulation, and shifts from earlier MRU preference
toward MRA later in the trace: early in the episode, reuse is favored because recently
accessed facts are still likely relevant, whereas later the newest observations more
often provide the decisive evidence for the current query. The trace therefore shows
adaptive switching among named symbolic QA heuristics as memory context changes: the
controller exposes changing preferences over named choices rather than a single opaque
action index. The corresponding exploration and forgetting traces are included in the
supplemental material.

\subsection{Final Memory State vs. Hidden-State Ground Truth}

\begin{figure}[tb]
\centering
\includegraphics[width=0.81\linewidth]{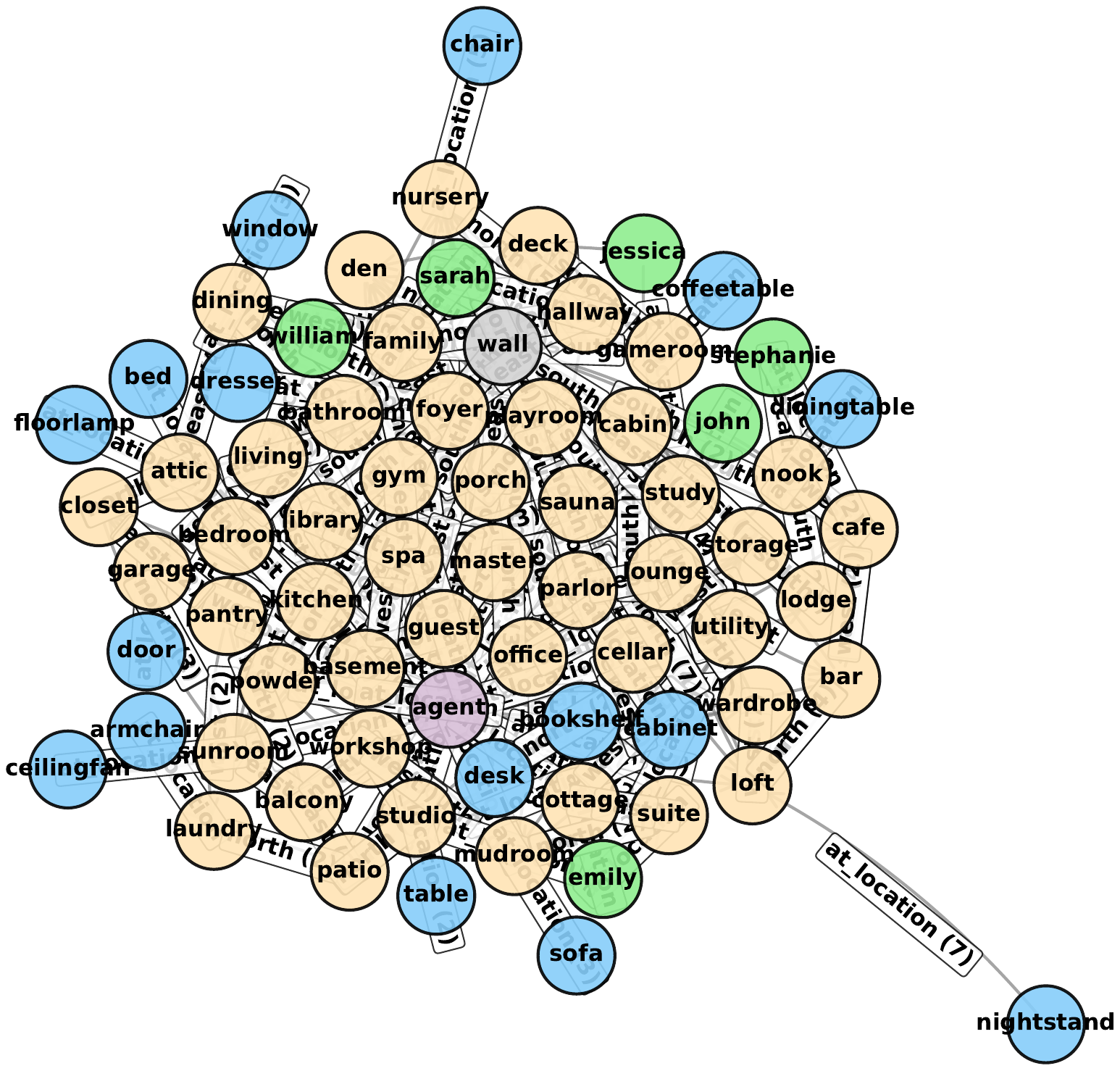}
\caption{Memory-state graph at the end of the held-out test trace. For visibility, edge
labels use parentheses $(N)$, where $N$ is the number of memories associated with the
same RDF triple. Because one RDF triple can correspond to multiple annotated statements,
these are condensed into a single displayed triple with count $(N)$ rather than listing
all annotations individually. This figure makes visible what the bounded long-term
memory actually retains, which is central to comparing learned heuristic preferences
with their resulting memory state. The supplement includes the full figure and memory
evolution from steps 0 to 99.}
\label{fig:memory-state-final}
\end{figure}

Figure~\ref{fig:hidden-state-rdf-step99} is the ground-truth hidden graph at $t=99$;
Figure~\ref{fig:memory-state-final} is the agent's long-term memory at the same time.
Memory covers all 49 rooms, showing that exploration reached every area. Object facts
are selective: in this run the memory retains 16 static-object facts and 6 moving-object
facts, while the hidden state has 18 of each.

This gap is expected under partial observability and a capacity limit. Static facts are
persistent and can be refreshed when rooms are revisited. Moving objects change and are
harder to reconfirm, so some are displaced once memory fills. The final memory is
therefore a task-driven summary, not a full reconstruction of the hidden graph.

Together, Figures~\ref{fig:q-values-qa} and \ref{fig:memory-state-final} show
both policy-side preferences and the resulting memory contents, supporting the claim that
the controller remains inspectable without relying on latent internal states.

\FloatBarrier
\section{Related Work}

\paragraph{Memory for partial observability in RL.}
Deep RL under partial observability is commonly addressed with recurrent or external
memory mechanisms, including DRQN-style recurrence and differentiable memory systems
\cite{mnih2013playingatarideepreinforcement,hausknecht2017deeprecurrentqlearningpartially,graves2014neuralturingmachines,Graves2016HybridCU,parisotto2017neuralmapstructuredmemory,pritzel2017neuralepisodiccontrol,wayne2018unsupervisedpredictivememorygoaldirected,Vincent_2018}.
These approaches improve long-horizon control, but the dominant memory state is
typically latent, making policy-level diagnosis difficult when one needs to explain
what was retained, forgotten, or retrieved at each step. Related analyses also highlight
generalization and bias issues under partial observability in batch RL settings
\cite{francoislavet2019overfittingasymptoticbiasbatch}.
Our design keeps the memory symbolic and the control choices named.

\paragraph{Relational and knowledge-graph representations.}
Relational inductive biases improve decision making in structured environments
\cite{zambaldi2018relationaldeepreinforcementlearning,schlichtkrull2017modelingrelationaldatagraph},
and qualifier-aware message passing further extends this to hyper-relational/
qualified knowledge graphs \cite{galkin2020messagepassinghyperrelationalknowledge}.
Broad knowledge-graph overviews provide the larger representational context
\cite{Hogan_2021}.
In parallel, RDF and RDF triple-annotation formalisms provide explicit graph semantics and
annotation-compatible representations for temporal/provenance-style metadata
\cite{cyganiak2014rdf11concepts,hayes2014rdf11semantics,rdf12concepts,rdf12turtle,sparql12query}. Our work
builds directly on this combination by learning over annotated RDF memory while
keeping executed memory operations symbolic. For neural KG query reasoning, related
lines include complex query answering and hyper-relational query embedding
approaches \cite{arakelyan2021complexqueryansweringneural,alivanistos2022query}.
Earlier temporal-RDF work studied validity-time semantics and querying directly in
RDF/OWL \cite{gutierrez2005temporalrdf,motik2012validitytime}. We position our use of
annotations differently: they serve as an implementation substrate for explicit
agent-memory metadata rather than as a new proposal for temporal RDF semantics or
query languages. We use qualifier-aware encoding to select symbolic memory operators
online, not to answer queries over a fixed graph. The encoder therefore serves
control rather than retrieval, aligning its signals with the recency/frequency/timestamp heuristics scored by the meta-policy.

\paragraph{Symbolic and neuro-symbolic control.}
Prior symbolic RL and neuro-symbolic RL studies emphasize that explicit symbolic
structure can improve interpretability and generalization
\cite{garcez2018symbolicreinforcementlearningcommon,ijcai2022p742,pmlr-v139-landajuela21a,harini2023neurosymbolicmetareinforcementlearning}.
For RoomKG specifically, Kim et al.
\cite{kim2026temporalknowledgegraphmemorypartially} introduced temporal KG memory and
established strong symbolic and neural reference points. In this paper, we evaluate
against representative in-domain baselines from both sides:
fixed symbolic heuristics and end-to-end neural sequence policies (LSTM/Transformer), in addition to our neuro-symbolic variants.
Our contribution lies in learning which symbolic heuristic to activate at each step
while preserving symbolic execution.

\paragraph{Hierarchical and meta-policy perspectives.}
Our formulation is also related to temporal abstraction and option selection
\cite{SUTTON1999181,bacon2016optioncriticarchitecture}. The key distinction is that
our selectable units are explicit symbolic heuristics (for QA, exploration, and
forgetting), not latent option embeddings.

\section{Conclusion}

We presented a neuro-symbolic meta-policy for partially observable
control with temporal knowledge-graph memory. The central idea is to learn
\emph{which} symbolic heuristic to apply (QA, exploration, forgetting) while keeping
execution symbolic and inspectable. In RoomKG at a memory capacity of 512, this improves over strong
symbolic and neural references; qualifier-aware StarE-GNN gives the best held-out test
performance while preserving step-level traceability.
The main result is that adaptivity and inspectability need not be traded off here:
learned meta-selection improves held-out performance while preserving symbolic execution
over explicit RDF memory.

The scope of our claims is this bounded-memory RoomKG setting. At the design level, the
method assumes only that (i) memory is an annotated RDF graph and (ii) memory operations
are expressible as named heuristics ranked over annotation values such as recency and
frequency; neither assumption is RoomKG-specific, so the formulation transfers in
principle to other temporal-knowledge-graph and memory-management settings.
Scalability to larger knowledge graphs is bounded by design: the controller only ever
encodes its bounded memory ($\leq K$ facts) and the room-local observation, never the
full world graph, so per-step cost does not grow with world size, though richer worlds
may warrant a larger $K$. Limitations remain: evidence covers one benchmark family and
one primary capacity (512); deterministic dynamics do not cover stochastic or
non-stationary regimes; and performance depends on the heuristic library. The spread
among the fixed baselines in Table~\ref{tab:main-results} shows that heuristic choice
matters, and a poor library would cap what learned selection can achieve. Future work
should broaden environments and capacity regimes, test stochastic robustness, and
expand heuristic libraries while retaining explicit decision traces.

\begin{credits}
\subsubsection{\ackname}
This research was (partially) funded by the Hybrid Intelligence Center, a 10-year
program funded by the Dutch Ministry of Education, Culture and Science through the
Netherlands Organization for Scientific Research,
\url{https://www.hybrid-intelligence-centre.nl/}.
\end{credits}

\section*{Supplemental Material Statement}
Source code and the full supplemental material are openly available at
\url{https://github.com/humemai/roomkg-meta-policies} (release \texttt{v0.1.0}). The
repository includes the training, evaluation, and plotting code; reproducibility
documentation with the complete hyperparameter settings; and the full-size/original
figures referenced in the paper, including the diagnostic Q-value plots for all three
heads and the memory-evolution sequence from steps 0 to 99. The main paper remains
self-contained for understanding the method and assessing the core claims.

\section*{Declaration of use of Generative AI}
GPT-5.4 (OpenAI) and Fable 5 (Anthropic) were used to polish and revise the text and
figures of this manuscript. All research, code, experiments, and writing were carried
out by the authors.

\bibliographystyle{splncs04}
\bibliography{references}
\end{document}